\def\year{2018}\relax
\documentclass[letterpaper]{article} 
\usepackage{aaai17}  
\usepackage{times}  
\usepackage{helvet}  
\usepackage{courier}  
\usepackage{url}  
\usepackage{graphicx}  
\usepackage{algorithm}
\usepackage{algorithmic}
\nocopyright
\graphicspath{ {imgs/} }

\title{\LARGE \bf Deep HMResNet Model for Human Activity-Aware Robotic Systems}

\author{Hazem Abdelkawy, Naouel Ayari, Abdelghani Chibani, Yacine Amirat and Ferhat Attal\\
LISSI Laboratory\\ 
University of Paris-Est Cr\'eteil (UPEC),\\
Vitry-sur-Seine France\\
\{hazem-khaled-mohamed.abdelkawy,naouel.ayari,abdelghani.chibani,
amirat,ferhat.attal\}@u-pec.fr}
\begin{document}
\maketitle
\thispagestyle{empty}
%
\pagestyle{empty}
\begin{abstract}
Endowing the robotic systems with cognitive capabilities for recognizing daily activities of humans is an important challenge, which requires sophisticated and novel approaches. Most of the proposed approaches explore pattern recognition techniques which are generally based on hand-crafted features or learned features. In this paper, a novel Hierarchal Multichannel Deep Residual Network (HMResNet) model is proposed for robotic systems to recognize daily human activities in the ambient environments. The introduced model is comprised of multilevel fusion layers. The proposed Multichannel 1D Deep Residual Network model is, at the features level, combined with a Bottleneck MLP neural network to automatically extract robust features regardless of the hardware configuration and, at the decision level, is fully connected with an MLP neural network to recognize daily human activities. Empirical experiments on real-world datasets and an online demonstration are used for validating the proposed model. Results demonstrated that the proposed model outperforms the baseline models in daily human activity recognition.
\end{abstract}
\section{Introduction}

There is a growing consensus about the need of adding some cognitive capabilities to the connected things and robots that are produced today in order to provide the added value of assistive services for dependent people. These services aim to improve their quality of life and their physical and mental well-being and to guarantee their safety \cite{SEBBAK2013}. In the context of ambient assisted living (AAL), daily human activity recognition (DHAR) is one of the interesting cognitive capabilities that must be present in any robotic system. In this context, some studies based on a new generation of data-driven approaches, deep learning models, were proposed recently in the literature \cite{plotz2011feature,yang2015deep,ronao2016human}.   

The development of robotic systems with the capability of daily human activity recognition requires sophisticated and novel approaches. To enable an efficient recognition of daily activities in a dynamic environment, an exhaustive activities sensing with fusion techniques are required. Building efficient cognitive models for robotic systems requires a suitable architecture allowing the integration of heterogeneous sensors, objects, and robots. 

In this paper, A novel Hierarchal Multichannel Deep Residual Network (HMResNet) model is proposed for robotic systems to recognize daily human activities in the ambient environments. The introduced model is comprised of multilevel fusion layers combined with residual shortcut connections. The proposed Multichannel 1D Deep Residual Network model is, at the features level, combined with a Bottleneck MLP neural network to automatically extract features and, at the decision level, is fully connected with an MLP neural network to recognize daily human activities. The hierarchical architecture of the proposed model gives the advantage of extracting more complex features than the traditional plain deep learning models and facilitates the training of more sparse and deep networks than those used in the literature.

The paper is organized as follows: First, we review the related works concerning daily human activity recognition in the robotics field. Then, we describe the proposed Hierarchal Multichannel Deep Residual Network (HMResNet) model for human daily activities recognition. We evaluate the performances of the proposed approach with extensive experiments on real-world datasets besides a scenario of cognitive daily exercises coaching. This paper is concluded with a short review of the proposed model and a summary of the ongoing works.
\section{Related Work}
\label{relwork}
Within the objective of developing added value services to assist dependent people in their daily activities, as well as identifying these activities receives much focus in recent years \cite{kashiwabara2012teroos}. One of the most challenging tasks in human activity recognition is the extraction of remarkable features from the raw inertial data. Most of the existing works are based on handcrafted engineered features, which are known as shallow features. The most commonly used features for human activity recognition are hand-engineered features \cite{anguita2012human}, time domain features such as mean, median, variance, skewness, kurtosis and range \cite{attal2015physical} and frequency domain features such as temporal fast Fourier transform (tFFT) \cite{sharma2008high}, Discrete Fourier Transform (DFT) and Power Spectral Density (PSD) \cite{attal2015physical}. To identify different daily human activities, classification approaches, such as, Hidden Markov Models \cite{lee2011activity}, Artificial Neural Network \cite{mantyjarvi2001recognizing}, Support Vector Machine (SVM) \cite{he2009activity} and naive Bayes classifiers \cite{yang2010implementation} rely on the latter features. 



As evidenced, The traditional machine learning approaches have made significant progress in the last decade. However, these methods are based on heuristic hand-engineered features. Besides, the more amount of input data, the more inability of those approaches to come up with such relevant and consistent features. Moreover, the majority of the latter approaches depend on learning from static data, while the recognition of human activities in real life is based on data streaming from heterogeneous sensors that require durable online and incremental learning.

In recent years, a fast development of deep learning models has been started to compensate the drawbacks of the traditional data-driven approaches. 
One of the first attempts to recognize human basic activities with a deep learning model is proposed in \cite{plotz2011feature}. This study based on the creation of Restricted Boltzmann Machines (RBM) allows extracting features from accelerometer raw data. In \cite{duffner20143d}, a convolutional neural network (ConvNet) model is proposed for recognizing basic gestures from the accelerometer and gyroscope raw data. This ConvNet model outperformed the other state-of-the-art models in gesture recognition such as Dynamic Time Wrapping (DTW) and Hidden Markov Model (HMM). A hierarchical ConvNets model is proposed in \cite{yang2015deep} and benchmarked against many datasets of daily activities to show its performance compared to other states of the art baseline models. In \cite{ronao2016human}, ConvNet model is proposed to recognize Human Activities using smartphone sensors.

Compared to the traditional machine learning approaches, deep learning approaches depend on learned-features that can be extracted from the raw data automatically. These features are more relevant and complex than the hand-engineered features used with the traditional machine learning approaches. Besides, the nature of the deep learning models structure allows performing online and incremental learning. Despite the above-mentioned advantages, compared to computer vision and natural language processing tremendous development, there are only a few attempts \cite{zheng2016exploiting,yang2015deep,lee2017human} that tried to exploit the hierarchical deep learning models to classify 1D time series which is the cornerstone of HAR.

\section{HMResNet Deep Learning Model}
\label{HMResNet}
In this paper, to recognize human daily activities in a dynamic environment, a new deep learning architecture based on Hierarchical Multichannel deep Residual Network (HMResNet) is proposed. 
Compared to the state of the art deep learning models \cite{zheng2016exploiting,ronao2016human,yang2015deep,lee2017human}, the proposed model is based on multilevel fusion layers, with residual shortcut connections, and working on the multichannel raw data. At the features level fusion, a Multichannel 1D Deep Residual Network combined with a Bottleneck MLP neural network is proposed, for each sensor channel, to automatically extract features from raw data. At the decision level fusion, a multi-sensor fusion layer based on deep 1D ResNet followed by a fully connected MLP neural network is exploited for recognizing daily human activities. Indeed, the hierarchical architecture of the proposed model gives the advantage of extracting more relevant and complex features than the traditional plain deep learning models and facilitates the training of more sparse and deep networks than those used previously in the literature.

\subsection{Raw Data Preprocessing}
This step consists of data filtering, segmentation, and missing values replacement processes. In this paper, the Inertial Measurement Unit (IMU) sensors are used to measure linear and angular motion based on accelerometer and gyroscope raw data. Noise filters are used for preprocessing the sensors raw data (gyroscope and accelerometer). The sliding window algorithm is applied to segment the separated input signals into fixed-size windows. Finally, the preprocessed data are transferred to the feature extraction/fusion layers as a vector of sliding windows which contains accelerometer and gyroscope preprocessed raw data components, as shown in Fig.\ref{DetailedModel}. 

\subsection{Feature Level Fusion}
\begin{figure*}[!ht]
\centering 
\includegraphics[scale=0.43]{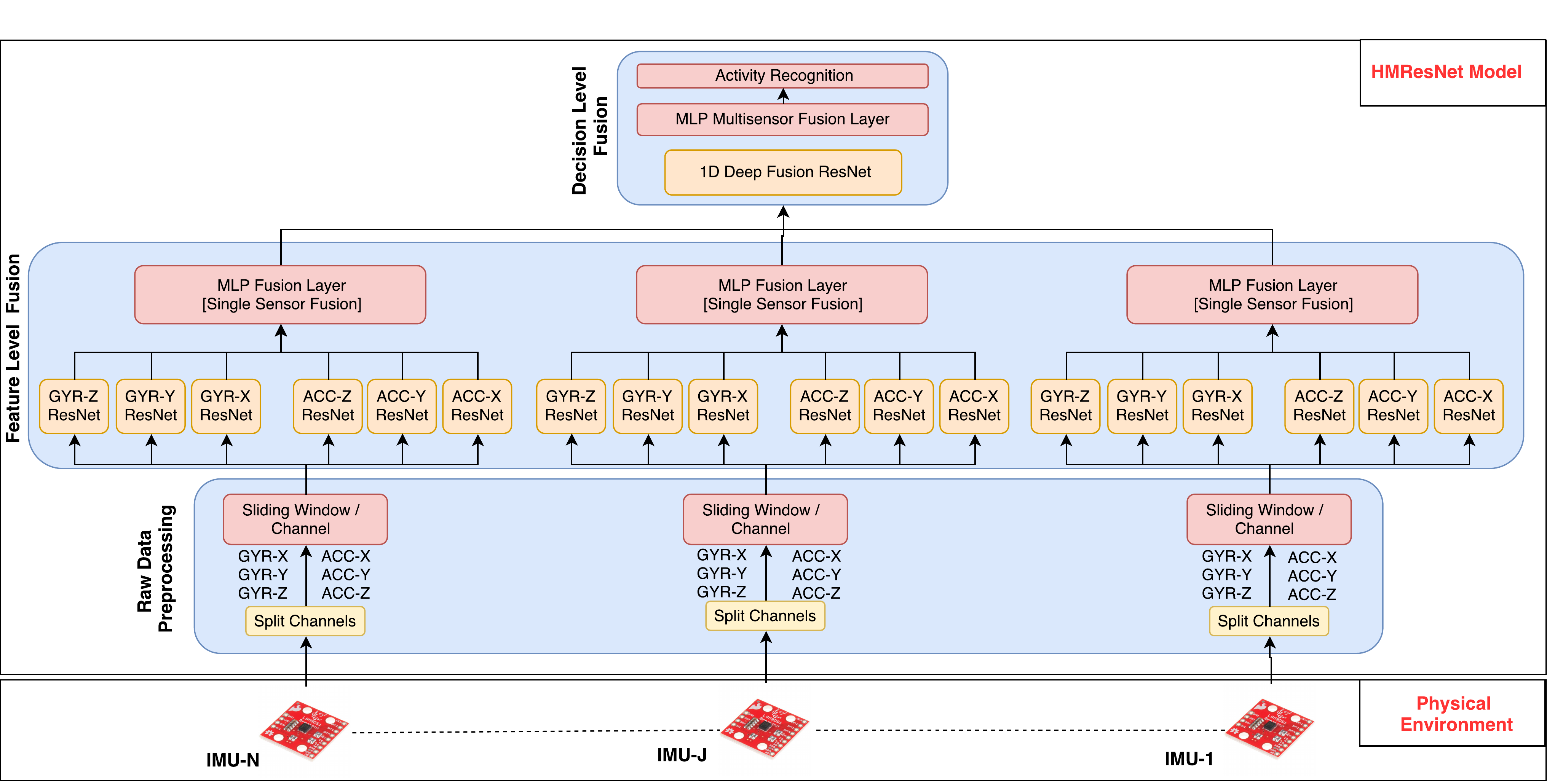}
\caption{Daily Human Activity Recognition based-Hierarchal Multichannel Deep Residual Network Model for robotic systems Exploiting $N$ IMUs}
\label{DetailedModel}
\end{figure*}

At the feature level fusion, the 1D deep ResNet is exploited to extract features automatically from the preprocessed raw data, followed by a Bottleneck MLP neural network to create sensor level features fusion layer as shown in Fig.\ref{DetailedModel}.

\subsubsection{Deep Residual Network (ResNet)}
Basically, the Deep Residual Network $(ResNet)$ developed by Microsoft research labs, is exploited in \cite{he2016deep} for image recognition. ResNet got the first place in the five main tracks of COCO and ImageNet competitions, which covering object recognition, image classification, and semantic segmentation. Hence, many studies started to evaluate ResNet performance in different fields such as speech recognition \cite{xiong2017microsoft}, and question answering systems \cite{de2017modulating}. However, to our knowledge, a single attempt was proposed in \cite{wang2017time} to use ResNet for time series classification. 

\begin{figure}
\centering 
\includegraphics[scale=0.36]{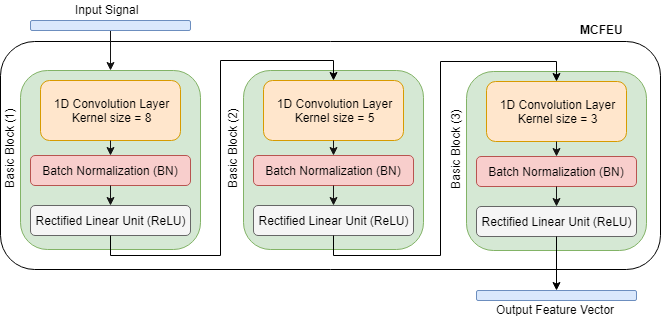}
\caption{Multilayer Convolution Feature Extractor Unit (MCFEU)}
\label{BasicBlock_Unit}
\end{figure}

\begin{figure}
\centering 
\includegraphics[scale=0.29]{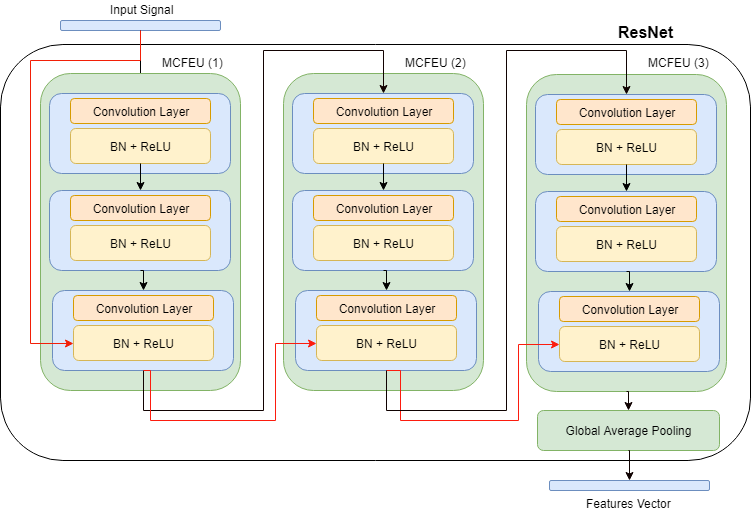}
\caption{Deep Residual Network Based on Stacked MCFEU units}
\label{ResnetNetwork}
\end{figure}

In the proposed model, the basic block of ResNet is $1D$ convolutional layer with kernel $(W_n)$ of size $(s)$ followed by Batch Normalization $(BN)$ \cite{DBLP:journals/corr/IoffeS15} and Rectified Linear Unit $(ReLU)$ layers. To avoid the problem of the vanishing gradient, ReLU activation function is used. The Batch normalization (BN) is applied to speed up the model convergence and improve the model generalization. Inspired by ResNet152 deep model \cite{he2016deep}, A plain network based on $3$ basic blocks are developed with different 1D kernel sizes, without strides, with 32, 64, and 64 feature maps respectively to create Multilayer Convolution Feature Extractor Unit (MCFEU) as shown in Fig.\ref{BasicBlock_Unit}. Both The kernel sizes and the number of feature maps have been chosen based on the empirical experiments which were conducted on different datasets \cite{ronao2016human,attal2015physical}. MCFEU is exploited to extract multilevel features from time series preprocessed raw data. The complete deep ResNet model is developed by stacking multiple MCFEU units, besides adding residual shortcut connection between the MCFEU units as shown in Fig.\ref{ResnetNetwork}. The shortcut connections are exploited to ensure that every MCFEU unit is learning more meaningful features and to solve the problem of the vanishing gradient for deep networks \cite{he2016deep}. Before feeding the extracted features to the decision level fusion layer, a Global Average Pooling $(GAP)$ layer \cite{lin2013network} is used to minimize the model overfitting by reducing the total number of the learned parameters.

\subsubsection{MLP Neural Network for Sensor Level Fusion}
In this paper, for every single IMU sensor, a bottleneck MLP neural network is exploited as a fusion layer for the sensor's channels as shown in Fig.\ref{DetailedModel}. The bottleneck MLP neural network acts as a nonlinear dimension reduction module, which is used for extracting low-dimensional features from the integrated deep ResNet output features. Finally, the outputs of all bottleneck MLP neural networks are integrated into a single feature vector, which is fed to the decision level fusion layer, see Algorithm \ref{MLPSensorFusion}. In this work, the bottleneck MLP neural network is comprised of : 
\begin{itemize}
  \item Two fully connected hidden layers where each layer consists of 1000 nodes.
  \begin{itemize}
  \item Each layer is denoted by $l_{i}$, where $i$ is the layer index.
  \item Each node is denoted by $n_{i}^{j}$, where $i$ is the layer index and $j$ is the node index inside a layer.
  \end{itemize}
  \item The Dropout algorithm \cite{srivastava2014dropout} is applied on 30\% of the nodes to prevent the network from overfitting the training features.
  \item To break the symmetry of the neurons performance,  The network weights are randomly initialized with small values close to zero based on normal distribution with $\mu = 0$ and $\sigma ^{2} = 0.05$ .
  
  \begin{itemize}
  \item Each weight is denoted by $w_{i,j}^{k}$, which refers to the weight from the node $n_{i}^{j}$ to the node $n_{i+1}^{k}$.
  \end{itemize}
  \item For every node, the Rectified Linear Unit (ReLU) is exploited as an activation function.
  \begin{itemize}
  \item Each node activation is denoted by $a_{i,j}$.
  \end{itemize}
\end{itemize}

\begin{algorithm}
\caption{Bottleneck MLP neural network Sensor Fusion Layer (BMLP)}
\label{MLPSensorFusion}
\hspace*{\algorithmicindent} \textbf{Input: X $\rightarrow$ List of feature vectors with length $n$} \\
\hspace*{\algorithmicindent} \textbf{Parameters to be learned: $w_{i,j}^{k}$, $b_j$} \\
\hspace*{\algorithmicindent} \textbf{Output: $\hat{F} = BMLP(X)$}
\begin{algorithmic}[1] 
\STATE \textbf{Randomly initialize the network weights}
\STATE $\hat{X} \leftarrow \textbf{Flatten the input features vector}$
\FOR{ each $n_{1}^{j}$}
\STATE $a_{1,k} \leftarrow RelU(\sum_{j}^{ } w_{1,j}^{k}.\hat{X}[j] + b_j)$
\STATE $A_{1}$.append($a_{1,k}$)
\ENDFOR \\
\STATE $A_{1} \leftarrow$ Dropout($A_{1}$,0.3)
\FOR{ each $n_{2}^{j}$}
\STATE $a_{2,k} \leftarrow RelU(\sum_{j}^{ } w_{2,j}^{k}.A_{1}[j] + b_j)$
\STATE $A_{2}$.append($a_{1,k}$)
\ENDFOR \\
\STATE $A_{2} \leftarrow$ Dropout($A_{2}$,0.3)
\STATE $\hat{F} \leftarrow flatten(A_{2})$ \\
\textbf{Return $\hat{F}$}
\end{algorithmic}
\end{algorithm}

\subsection{Decision level fusion}
At decision level fusion, both the 1D deep ResNet and the Bottleneck MLP neural networks are exploited to recognize daily human activities as shown in Fig.\ref{DetailedModel}. The Bottleneck MLP neural network is comprised of three fully connected layers. In the hidden layers, each layer consists of 1000 nodes which are based on ReLU activation function. The output layer consists of a number of nodes equal to the total number of target activities. Besides, the softmax activation function is used for the output layer, see Algorithm \ref{DecisionFusion}. During the training phase, the categorical cross entropy cost function is exploited to calculate the difference between the target labels and the predicted labels. This difference is exploited by the backpropagation algorithm \cite{hagan1994training} to update the parameters to be learned of both the feature level and decision level layers during the training phase. Finally, Adam algorithm \cite{kingma2014adam} is used for optimizing the MLP categorical cross-entropy cost function.

\begin{algorithm}
\caption{Decision Level Fusion Classifier}
\label{DecisionFusion}
\hspace*{\algorithmicindent} \textbf{Input: X $\rightarrow$ 1D Feature vector with length $n$} \\
\hspace*{\algorithmicindent} \textbf{Parameters to be learned: $w_{i,j}^{k}$, $b_j$} \\
\hspace*{\algorithmicindent} \textbf{Output: $\hat{Y} = Predict(X)$}
\begin{algorithmic}[1] 
\STATE $\hat{X} \leftarrow ResNet(X) $
\FOR{ each $n_{1}^{j}$}
\STATE $a_{1,k} \leftarrow RelU(\sum_{j}^{ } w_{1,j}^{k}.\hat{X}[j] + b_j)$
\STATE $A_{1}$.append($a_{1,k}$)
\ENDFOR \\
\STATE $A_{1} \leftarrow$ Dropout($A_{1}$,0.3)
\FOR{ each $n_{2}^{j}$}
\STATE $a_{2,k} \leftarrow RelU(\sum_{j}^{ } w_{2,j}^{k}.A_{1}[j] + b_j)$
\STATE $A_{2}$.append($a_{1,k}$)
\ENDFOR \\
\STATE $A_{2} \leftarrow$ Dropout($A_{2}$,0.3)
\STATE $\hat{F} \leftarrow flatten(A_{2})$ \\
\FOR{ each $n_{3}^{j}$}
\STATE $a_{3,k} \leftarrow Softmax(\sum_{j}^{ } w_{3,j}^{k}.A_{2}[j] + b_j)$
\STATE $A_{3}$.append($a_{3,k}$)
\ENDFOR \\
\STATE $\hat{Y} \leftarrow A_{3}.getIndex(max(A_{3}))$) \\
\textbf{Return $\hat{Y}$}
\end{algorithmic}
\end{algorithm}


\section{Experiments and Evaluation}
\label{exp}

The proposed approach for daily human activities recognition is evaluated through empirical experiments on real-world datasets: Smartphones dataset \cite{anguita2013public} and Wearable Sensors dataset \cite{attal2015physical}. The proposed model is evaluated against the following baseline models : (i) k-NN with time domain and frequency domain features \cite{attal2015physical} using the Wearable Sensors dataset. (ii) Convnet combined with MLP neural network applied to raw data \cite{ronao2016human}, and Convnet with tFFT features \cite{ronao2016human}, using the Smartphones dataset. The architecture shown in Fig.\ref{DetailedModel} is used for both datasets, the only differences are the number of input sensors and the number of output classes since the input and output of the datasets are different. Both of the features level deep networks and the decision level deep networks are trained together to ensure the consistency of the learning process.  

\subsection{Datasets}
Basically, human activities are divided into periodic, static, or sporadic activities. Periodic activities such as biking and walking, static activities such as standing, lying and seated, and sporadic activities are intention-oriented activities such as drinking from a cup, and opening door \cite{bulling2014tutorial}.

\subsubsection{Smartphone Dataset}
In this dataset, the data were collected using a smartphone with a built-in accelerometer and gyroscope tri-axial sensor. The dataset consists of 6 different activities performed by 30 volunteer subjects while holding a smartphone in a pocket tight around their waist. The activities are a mix of periodic and static activities such as \textit{WALKING, WALKING UPSTAIRS, WALKING DOWNSTAIRS, SITTING, STANDING, and LAYING}. The data were sampled at 50Hz and divided into fixed length windows of 128 samples with 50\% overlap. Butterworth low-pass filter is used to separate body acceleration and gravity from the accelerometer raw data. The data were separated into a training set with 7352 windows from 21 randomly selected subjects, and testing set of the remaining 2947 windows.

\begin{table}
\centering
\caption{Smartphone dataset Accuracy Evaluation}
\label{SmartPhoneACC}
\begin{tabular}{|c|l|l|c|l|}
\hline
\multicolumn{3}{|c|}{Method}                               & \multicolumn{2}{c|}{Accuarcy (\%)}          \\ \hline
\multicolumn{5}{|c|}{\textbf{Baseline Models \cite{ronao2016human}}}                                                      \\ \hline
\multicolumn{3}{|c|}{PCA+MLP}                              & \multicolumn{2}{c|}{57.10}           \\ \hline
\multicolumn{3}{|c|}{HCF+NB}                               & \multicolumn{2}{c|}{74.32}           \\ \hline
\multicolumn{3}{|c|}{HCF+J48}                              & \multicolumn{2}{c|}{83.02}           \\ \hline
\multicolumn{3}{|c|}{SDAE+MLP(DBN)}                        & \multicolumn{2}{c|}{87.77}           \\ \hline
\multicolumn{3}{|c|}{HCF+ANN}                              & \multicolumn{2}{c|}{91.08}           \\ \hline
\multicolumn{3}{|c|}{HCF+SVM}                              & \multicolumn{2}{c|}{94.61}           \\ \hline
\multicolumn{5}{|c|}{\textbf{Deep Learning Models \cite{ronao2016human}}}                                                 \\ \hline
\multicolumn{3}{|c|}{Convnet (inverted pyramid archi)+MLP } & \multicolumn{2}{c|}{94.79}           \\ \hline
\multicolumn{3}{|c|}{tFFT+Convnet (($J(L_1)$)=200)  }        & \multicolumn{2}{c|}{95.75}           \\ \hline
\multicolumn{3}{|c|}{\textbf{Proposed Model}}              & \multicolumn{2}{l|}{}                  \\ \hline
\multicolumn{3}{|c|}{\textbf{Hierarchal Multichannel Deep ResNet}}    & \multicolumn{2}{c|}{\textbf{97.619}} \\ \hline
\end{tabular}
\end{table}

\begin{table*}[!ht]
\centering
 \caption{HAR Using Smartphone dataset Confusion Matrix Evaluation}
\label{SmartPhoneConfMat}
\begin{tabular}{ccccccclcccccc}
\multicolumn{1}{l}{} & \multicolumn{6}{c}{\textbf{ConvNet Predicted classes}} & & \multicolumn{6}{c}{\textbf{Proposed Model Predicted classes}}\\ 
\cline{2-7} \cline{9-14} 
\multicolumn{1}{c|}{\textbf{Actual class}}&\multicolumn{1}{c|}{\textbf{W}}& \multicolumn{1}{c|}{\textbf{WU}}&\multicolumn{1}{c|}{\textbf{WD}}  & \multicolumn{1}{c|}{\textbf{Si}}  & \multicolumn{1}{c|}{\textbf{St}}  & \multicolumn{1}{c|}{\textbf{L}}   & \multicolumn{1}{l|}{} & \multicolumn{1}{c|}{\textbf{W}}   & \multicolumn{1}{c|}{\textbf{WU}}  & \multicolumn{1}{c|}{\textbf{WD}}  & \multicolumn{1}{c|}{\textbf{Si}}  & \multicolumn{1}{c|}{\textbf{St}}  & \multicolumn{1}{c|}{\textbf{L}}   \\
\cline{1-7} \cline{9-14} 
\multicolumn{1}{|c|}{\textbf{Walking}}& \multicolumn{1}{c|}{\textbf{491}} & \multicolumn{1}{c|}{3}            & \multicolumn{1}{c|}{2}            & \multicolumn{1}{c|}{0}            & \multicolumn{1}{c|}{0}            & \multicolumn{1}{c|}{0}            & \multicolumn{1}{l|}{} & \multicolumn{1}{c|}{\textbf{496}} & \multicolumn{1}{c|}{0}            & \multicolumn{1}{c|}{0}            & \multicolumn{1}{c|}{0}            & \multicolumn{1}{c|}{0}            & \multicolumn{1}{c|}{0}            \\ 
\cline{1-7} \cline{9-14} 
\multicolumn{1}{|c|}{\textbf{W. upstairs}}   & \multicolumn{1}{c|}{0}            & \multicolumn{1}{c|}{\textbf{471}} & \multicolumn{1}{c|}{0}            & \multicolumn{1}{c|}{0}            & \multicolumn{1}{c|}{0}            & \multicolumn{1}{c|}{0}            & \multicolumn{1}{l|}{} & \multicolumn{1}{c|}{3}            & \multicolumn{1}{c|}{\textbf{468}} & \multicolumn{1}{c|}{0}            & \multicolumn{1}{c|}{0}            & \multicolumn{1}{c|}{0}            & \multicolumn{1}{c|}{0}            \\ 
\cline{1-7} \cline{9-14} 
\multicolumn{1}{|c|}{\textbf{W. downstairs}} & \multicolumn{1}{c|}{0}            & \multicolumn{1}{c|}{0}            & \multicolumn{1}{c|}{\textbf{420}} & \multicolumn{1}{c|}{0}            & \multicolumn{1}{c|}{0}            & \multicolumn{1}{c|}{0}            & \multicolumn{1}{l|}{} & \multicolumn{1}{c|}{2}            & \multicolumn{1}{c|}{0}            & \multicolumn{1}{c|}{\textbf{417}} & \multicolumn{1}{c|}{0}            & \multicolumn{1}{c|}{1}            & \multicolumn{1}{c|}{0}            \\ 
\cline{1-7} \cline{9-14} 
\multicolumn{1}{|c|}{\textbf{Sitting}}       & \multicolumn{1}{c|}{0}            & \multicolumn{1}{c|}{0}            & \multicolumn{1}{c|}{0}            & \multicolumn{1}{c|}{\textbf{436}} & \multicolumn{1}{c|}{34}           & \multicolumn{1}{c|}{21}           & \multicolumn{1}{l|}{} & \multicolumn{1}{c|}{0}            & \multicolumn{1}{c|}{1}            & \multicolumn{1}{c|}{0}            & \multicolumn{1}{c|}{\textbf{438}} & \multicolumn{1}{c|}{52}           & \multicolumn{1}{c|}{0}            \\
\cline{1-7} \cline{9-14} 
\multicolumn{1}{|c|}{\textbf{Standing}}      & \multicolumn{1}{c|}{0}            & \multicolumn{1}{c|}{1}            & \multicolumn{1}{c|}{0}            & \multicolumn{1}{c|}{24}           & \multicolumn{1}{c|}{\textbf{496}} & \multicolumn{1}{c|}{11}           & \multicolumn{1}{l|}{} & \multicolumn{1}{c|}{0}            & \multicolumn{1}{c|}{0}            & \multicolumn{1}{c|}{0}            & \multicolumn{1}{c|}{10}           & \multicolumn{1}{c|}{\textbf{522}} & \multicolumn{1}{c|}{0}            \\ 
\cline{1-7} \cline{9-14} 
\multicolumn{1}{|c|}{\textbf{Laying}}        & \multicolumn{1}{c|}{0}            & \multicolumn{1}{c|}{0}            & \multicolumn{1}{c|}{0}            & \multicolumn{1}{c|}{43}           & \multicolumn{1}{c|}{23}           & \multicolumn{1}{c|}{\textbf{471}} & \multicolumn{1}{l|}{} & \multicolumn{1}{c|}{0}            & \multicolumn{1}{c|}{0}            & \multicolumn{1}{c|}{0}            & \multicolumn{1}{c|}{0}            & \multicolumn{1}{c|}{0}            & \multicolumn{1}{c|}{\textbf{537}} \\ 
\cline{1-7} \cline{9-14} 
\end{tabular}
\end{table*}

\subsubsection{Wearable Sensors dataset}
In this dataset, the data were collected using three IMU sensors placed on the chest, the right thigh and the left ankle of the subject. Each IMU sensor has built-in tri-axial accelerometer, gyroscope, and magnetometer. The dataset consists of $12$ different activities performed by $6$ volunteer subjects. The activities are a mix of periodic and static activities with transitional activities such as \textit{A1: WALKING DOWNSTAIRS, A2: STANDING, A3: SITTING DOWN, A4: SITTING, A5: FROM SITTING TO SITTING ON THE GROUND, A6: SITTING ON THE GROUND, A7: LYING DOWN, A8: LYING, A9: FROM LYING TO SITTING ON THE GROUND, A10: STANDING UP, A11: WALKING, and A12: WALKING UPSTAIRS}. The dataset was sampled at $25$Hz and no sliding window was applied to the raw data.

\subsection{Baselines}
With regard to smartphone dataset, the baseline consists of two deep learning models which are applied to both the raw data and tFFT features extracted from a single smartphone IMU sensor \cite{ronao2016human}. The first baseline consists of ConvNet combined with MLP neural network to extract features automatically from the IMU sensor raw data. The second baseline consists of the ConvNet applied to tFFT features extracted from smartphone IMU sensor raw data. Both the latter baselines aim to recognize 6 static and periodic activities. The baseline models were evaluated over a test set of randomly selected 9 volunteers.

With regard to wearable sensors dataset, the baseline consists of multiple traditional machine learning models which are applied to both the raw data and time domain, frequency domain features extracted from 3 different IMU sensors \cite{attal2015physical}. The baseline models aim to recognize a set of static, periodic and transitional activities. The baseline models evaluated over 10 folds cross-validation with 25 samples and 80\% overlap sliding windows.

In this paper, To ensure a fair comparison with the baseline models, The same number of cross-folds, number of samples per sliding window, and sliding windows overlapping configurations are applied as the reference models.

\subsection{Results}

\begin{figure}
\centering 
\includegraphics[scale=0.43]{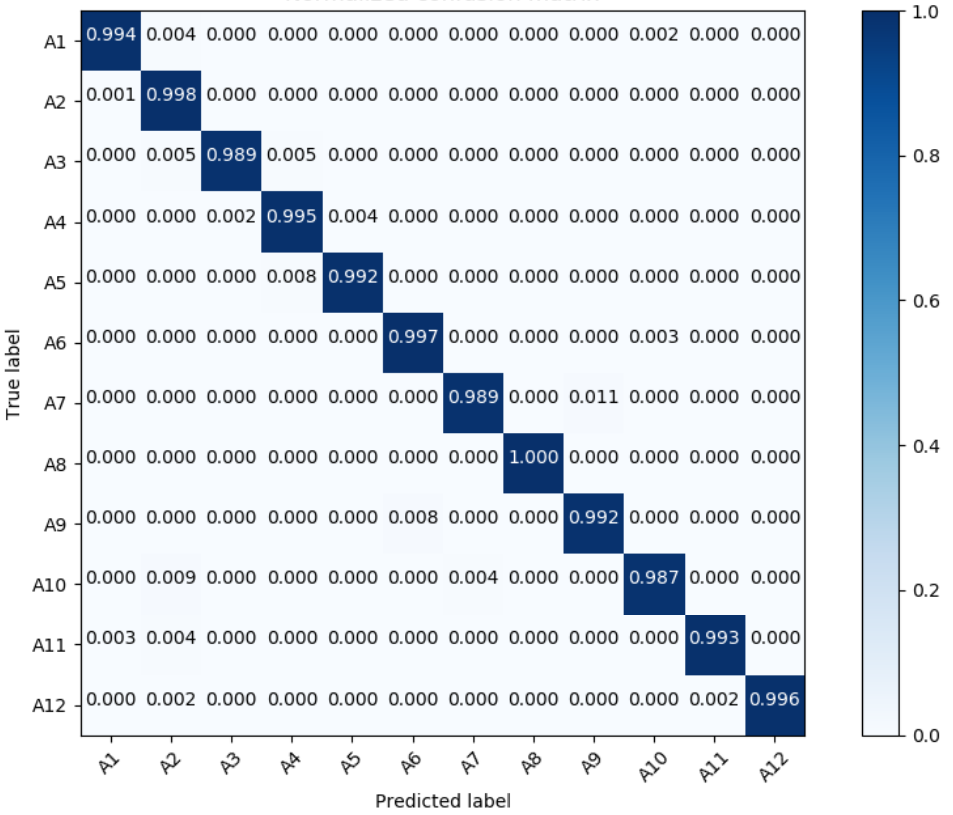}
\caption{Confusion Matrix obtained by HMResNet using wearable sensors dataset}
\label{confmatrix}
\end{figure}

With regard to smartphone dataset, In terms of accuracy, The results of the proposed model compared to eight baseline models are shown in Table \ref{SmartPhoneACC}. Besides, The confusion matrix of the best baseline model (ConvNet) compared to the proposed model (HMResNet) is shown in Table. \ref{SmartPhoneConfMat}. The best results are highlighted in bold for both tables.

From the Accuracy results, The proposed HMResNet model has significantly outperformed the baseline models and obtained better classification accuracy.
Compared to the best baseline model (ConvNet), The confusion matrix shows that a significant improvement can be observed for static activities such as $SITTING$, $STANDING$, and $LAYING$, which constituted a major impediment for the baseline model to classify them correctly. For $LAYING$ activity, the number of the correctly classified classes improved by 12.3\% with no misclassified classes. For $STANDING$ activity, the number of the correctly classified classes improved by 4.8\%, Besides the number of misclassified classes decreased by  4.5\%. For $SITTING$ activity, the number of the correctly classified classes improved by 0.4\%, Besides the number of misclassified classes decreased by 0.6\%.  
Based on the results, the proposed model shows that the hierarchal architecture with multilevel fusion layers combined with residual shortcut connections succeed to extract more relevant features for both static and periodic activities than both the hand engineered features and the plain CNN learned features. Besides, the proposed model is more accurate than both the traditional machine learning and deep learning baseline models for recognizing human daily activities based on a single accelerometer and gyroscope tri-axial sensor.

To evaluate the extendability and the robustness of the proposed model regardless of the hardware configuration, The proposed model was benchmarked against another baseline models using wearable sensors dataset. In terms of average precision-recall values, The results of the proposed model compared to eight base line models are shown in Table \ref{Werable}. 

With regard to baseline models, The K-nearest neighbor (k-NN) algorithm applied to time-domain, and frequency-domain features achieves the best results in terms of average recall, and precision values, followed by Random Forest (RF), then k-NN without features and finally the Supervised Learning Gaussian Mixture Models (SLGMM) without features obtains relatively the worst results. As shown in Table \ref{Werable}, the proposed model outperforms the baseline models which are evaluated on raw data as well as on the hand-crafted features. The obtained results show that the proposed model improved the values of precision-recall to be $99.22$\% and $98.88$\% respectively compared to the baselines methods that are varying from $69.88$\% to $98.85$\% and from $69.99$\% to $98.85$\%. The best results are highlighted in bold for in the table. The proposed model obtained almost perfect results as shown in the confusion matrix in Fig. \ref{confmatrix}.

Because of the small size of the dataset, the results show a slight difference when comparing the results of the proposed (HMResNet) model to the best baseline (k-NN with features) model. Even though the small difference, the extraction of features phase requires integrating additional models and algorithms to the baseline models. Besides, the feature extraction phase needs extra computation time, which is not practical for real-time applications. 

From the latter empirical evaluation experiments, The proposed HMResNet model succeeded to outperforms the baseline models, extract more relevant features, and recognize periodic, transitional, and static daily human activities from single IMU sensor up to $3$ IMU sensors, which shows the robustness of the proposed model regardless of the hardware configuration. 

\begin{table}[!ht]
\centering
\caption{Wearable Sensors dataset Evaluation}
\label{Werable}
\centering
\begin{tabular}{|l|c|c|l|}
\hline
\multicolumn{1}{|c|}{Model}                                  &Precision(\%) &Recall(\%)                          \\ \hline
\multicolumn{3}{|c|}{\textbf{Without features \cite{attal2015physical}}}            \\ \hline
\multicolumn{1}{|c|}{KNN}                           & 94.62                 & 94.57                      \\ \hline
\multicolumn{1}{|c|}{RF}                            & 83.46                 & 82.28                      \\ \hline
\multicolumn{1}{|c|}{SVM}                          & 90.33                 & 90.98                       \\ \hline
\multicolumn{1}{|c|}{SLGMM}                        & 69.88                 & 69.99                       \\ \hline
\multicolumn{3}{|c|}{\textbf{With features \cite{attal2015physical}}}               \\ \hline
\multicolumn{1}{|c|}{KNN}                          & 98.85                 & 98.85                        \\ \hline
\multicolumn{1}{|c|}{RF}                           & 98.25                 & 98.24                          \\ \hline
\multicolumn{1}{|c|}{SVM}                          & 92.90                 & 93.15                          \\ \hline
\multicolumn{1}{|c|}{SLGMM}                        & 73.61                 & 74.44                         \\ \hline
\multicolumn{3}{|c|}{\textbf{Proposed Model}}                                                         \\ \hline
\textbf{Deep Multichannel ResNet} & \textbf{99.22}                 & \textbf{98.88}  \\ \hline
\end{tabular}
\end{table}

\subsection{Real World Use case: Cognitive daily exercises coaching}

\begin{figure}[!ht]
\centering 
\includegraphics[scale=0.3]{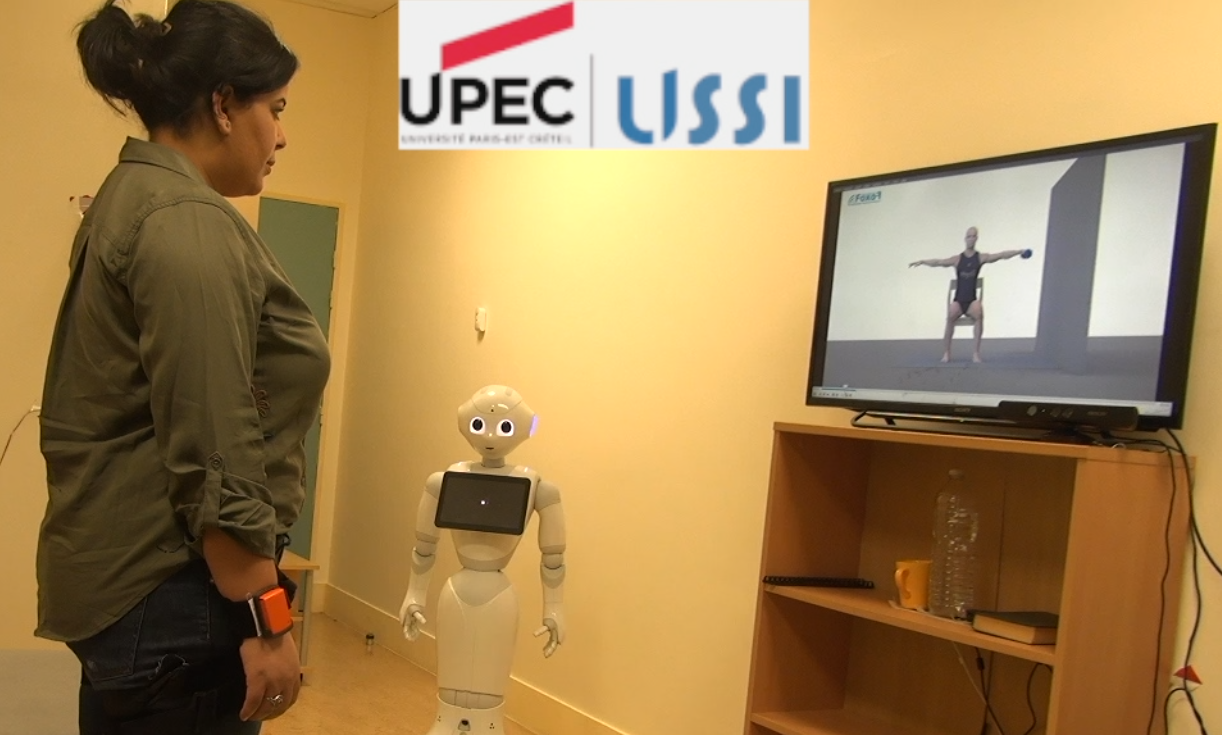}
\caption{Scene extracted from the smart home environment} 
\label{showroom}
\end{figure}

To validate the proposed approach for real-time activity recognition, a use case of cognitive daily exercises coaching for a diabetic person is studied, see Fig.\ref{showroom}. This use case consists of a robot, called Pepper, that is acting as a training coach of a diabetic person, called Alice. Indeed, Pepper recognizes and guide the daily exercises which were prescribed by a doctor for \textit{Alice}. This work is reported in a multimedia video that is available on \textit{LISSI}'s Website \footnote{\url{http://www.lissi.fr/videos/HMResNet.php}}.

\begin{figure}
\centering 
\includegraphics[scale=0.45]{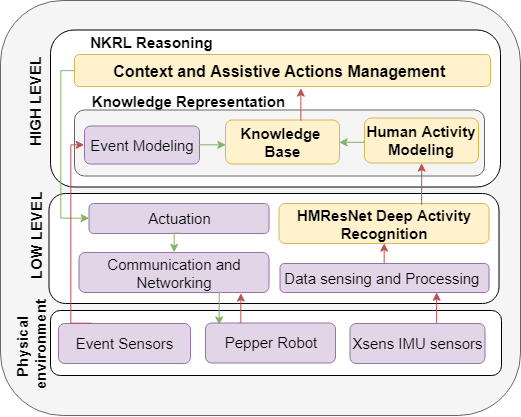}
\caption{Cognitive Architecture for Human Activities Aware Robotic Systems}
\label{arch}
\end{figure}

By extending a previously proposed cognitive architecture \cite{ayari17a}, Pepper can detect and monitor Alice's activities continuously based on the proposed HMResNet model. During this use case, a set of complex activities are recognized and analyzed by integrating the proposed HMResNet approach with previously proposed Narrative Knowledge Representation Language (NKRL) reasoning rules \cite{ayari17a,ayari17b,ayari16},  see Fig.\ref{arch}. 

At the low level, a \textit{communication service} is implemented to enable the entities populating the ambient environment to connect and subscribe to cloud services as well as to interchange knowledge. The \textit{communication service} is based on standard communication technologies such as (XMPP, REST, etc.). In addition to the \textit{communication service}, HMResNet Activity recognition, and Multi-modal data sensing services are implemented at the low level. 

At the high level, Knowledge representation services are proposed to model the dynamic knowledge. The knowledge representation services exploited the narrative knowledge representation language (NKRL) based n-ary ontologies, to avoid the problems experienced by binary ontologies for dynamic knowledge representation \cite{Tenorth2015,Lemaignan2016}. The representation services create NKRL predicates occurrences and save them in a shared knowledge base, thus the reasoning model query the predicates occurrences back when necessary.

During the experiment, The proposed model was evaluated by streaming 6974 sliding windows of 1.5 seconds and the average processing time to recognize a single activity was $0.2$ seconds. Therefore, The processing time is reasonably fitting the constraints of real-time activity recognition. 

This use case is a part of $MEDOLUTION$ European project \footnote{https://itea3.org/project/medolution.html} which is funded by $ITEA3$ Research, Development and Innovation (RDI) program.

\section{Conclusion}
\label{conc}

In this paper, a new deep learning architecture based on Hierarchical Multichannel deep Residual Network (HMResNet) is proposed for robotic systems to recognize daily human activities in ambient environments. The introduced model consists of multilevel fusion layers. The proposed Multichannel 1D Deep Residual Network model is, at features level, combined with Bottleneck MLP neural network to automatically extract relevant features and, at decision level, fully connected with MLP neural network to recognize daily human activities. 

The performance of the daily human activity recognition based on HMResNet model is shown through two datasets. The proposed automatic features extraction model is more relevant than both the hand engineered features and the plain CNN learned features. It is able to recognize perfectly, in terms of precision, the static activities: \textit{SITTING, STANDING,}  and \textit{LAYING}. In general, results demonstrated that the proposed model outperforms baseline methods exploiting the same datasets. 

To validate the proposed approach for real time activity recognition, a use case of daily exercises coaching for a diabetic person is studied.

The ongoing works address the extension of the proposed approach to explain human behavior through recognized activities over time. Besides, The further study of the hyper-parameters tunning,  and the extracted features by HMResNet model to evaluate them against the well-known baseline deep learning models should be addressed. Even though HMResNet neural networks can be a cornerstone technique for the
Human Activity Recognition, further study of the method and evaluating
larger datasets should be conducted.
\bibliographystyle{aaai}
\bibliography{AAAI}
\end{document}